\documentclass[runningheads]{llncs}
\usepackage[T1]{fontenc}
\usepackage{graphicx}
\usepackage{booktabs}
\usepackage{multirow}
\usepackage{afterpage}
\usepackage{caption}
\usepackage{amssymb}
\usepackage{hyperref}
\usepackage[table,xcdraw]{xcolor}
\begin{document}
\title{NT-VOT211: A Large-Scale Benchmark for Night-time Visual Object Tracking}
\titlerunning{NT-VOT211: A Benchmark for Night-time VOT }
%
\author{Yu Liu\inst{1}\orcidID{0009-0009-5898-0113} \and
Arif Mahmood\inst{2}\orcidID{0000-0001-5986-9876} \and
Muhammad Haris Khan\inst{3}\orcidID{0000-0001-9746-276X} \\ \textcolor{blue}{\textbf{Oral Acceptance at the Asian Conference on Computer Vision (ACCV) 2024, Hanoi, Vietnam.}}}
\authorrunning{Yu Liu et al.}
%
\institute{Xinjiang University\\
\email{750184785ly@gmail.com}
\and
Information Technology University\\
\email{arif.mahmood@itu.edu.pk}
\and
Mohamed bin Zayed University of Artificial Intelligence\\
\email{muhammad.haris@mbzuai.ac.ae}
}
\maketitle              
\begin{abstract}
Many current visual object tracking benchmarks such as OTB100, NfS, UAV123, LaSOT, and GOT-10K, predominantly contain day-time scenarios while the challenges posed by the night-time has been less investigated. It is primarily because of the lack of a large-scale, well-annotated night-time benchmark for rigorously evaluating tracking algorithms. To this end, this paper presents NT-VOT211, a new benchmark tailored for evaluating visual object tracking algorithms in the challenging night-time conditions.
NT-VOT211 consists of 211 diverse videos, offering 211,000 well-annotated frames with 8 attributes including camera motion, deformation, fast motion, motion blur, tiny target, distractors, occlusion and out-of-view. To the best of our knowledge, it is the largest night-time tracking benchmark to-date that is specifically designed to address unique challenges such as adverse visibility, image blur, and distractors inherent to night-time tracking scenarios. Through a comprehensive analysis of results obtained from 42 diverse tracking algorithms on NT-VOT211, we uncover the strengths and limitations of these algorithms, highlighting opportunities for enhancements in visual object tracking, particularly in environments with suboptimal lighting. Besides, a leaderboard  for revealing performance rankings, annotation tools, comprehensive meta-information and all the necessary code for reproducibility of results is made publicly available. We believe that our NT-VOT211 benchmark will not only be instrumental in facilitating field deployment of VOT algorithms, but will also help VOT enhancements and it will unlock new real-world tracking applications. Our dataset and other assets can be found at: \href{https://github.com/LiuYuML/NV-VOT211}{https://github.com/LiuYuML/NV-VOT211}

\end{abstract}
\section{Introduction}
\label{sec:intro}

\begin{figure*}[htbp]
\centering
\includegraphics[scale=0.35]{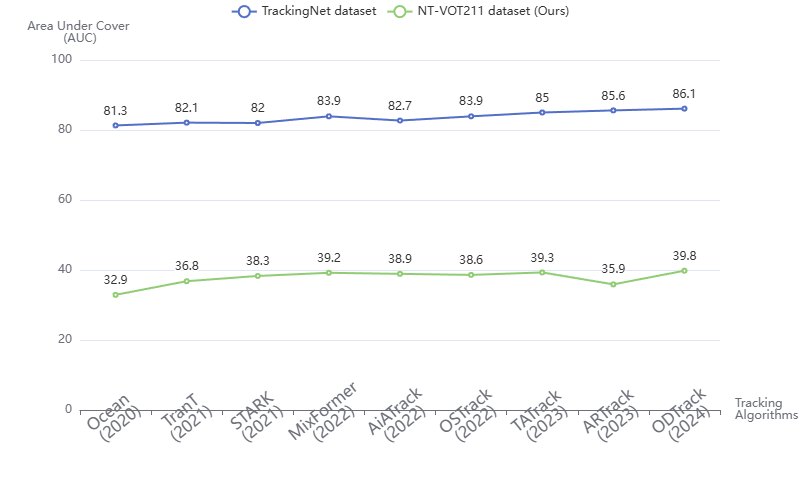}
\caption{Performance comparison of 9 SOTA VOT algorithms on a dominantly day-time dataset (TrackingNet) and on our proposed night-time dataset NT-VOT211. The performance of each algorithm is \emph{significantly poor} on our night-time benchmark underscoring the pressing need for night-time VOT benchmarks to enable reliable deployment of VOT algorithms in night-time scenarios. 
}
\label{yearly_result}
\end{figure*}


In recent years, the field of Visual Object Tracking (VOT) has undergone substantial progress, characterized by the proliferation of diverse tracking algorithms \cite{Bertinetto_2016_CVPR,bhat2020learning,cui2022mixformer,chen2023seqtrack,fu2021stmtrack,gao2022aiatrack}. This advancement has been fueled by the creation of several benchmark datasets, each playing a crucial role in the evaluation and comparison of tracking methodologies. Prominent among these datasets are OTB100 \cite{7001050}, NfS \cite{kiani2017need}, GOT-10k \cite{huang2019got}, TNL2K\cite{wang2021towards}, TrackingNet \cite{muller2018trackingnet}, LaSOT \cite{fan2019lasot},  AVisT \cite{noman2022avist}, and   the VOT competition \cite{VOT_TPAMI}. However,  all of these datasets lack the very important tracking attributes required for improving night-time VOT. Despite the fact that   any deployed VOT system has to handle night-time scenarios almost 50\% of the time, which is quite crucial for safety and security; existing VOT benchmarks significantly lack this aspect.


VOT is a crucial task with applications in security surveillance, autonomous driving, and wildlife conservation. Despite the significance of these applications, existing tracking algorithms have not effectively addressed the challenges posed by night-time conditions, which are prevalent in real-world scenarios. It is mainly because the current datasets fall short of significant amount of night-time scenarios for benchmarking SOTA trackers.  This could compromise the effectiveness of the tracking process during night-time. Figure \ref{yearly_result} shows a comparison of performance degradation of many SOTA trackers when applied to the proposed night-time tracking dataset compared to a popular existing day-time benchmark.


The research progress towards night-time robust, practical, and accurate visual object tracking has been significantly hindered by the absence of any well-annotated  night-time dataset and a supportive community. 
There is a pressing need to develop new datasets for the training of tracking algorithms capable of understanding the state of a target under various night-time conditions. To address the existing gap, we have developed a novel benchmark NT-VOT211. It is a comprehensive benchmark characterized by its diverse content, distinctive low-brightness attributes, and large size.
Existing night-time datasets including AVisT\cite{noman2022avist}, UAVDark135\cite{UAVDark}, NAT2021\cite{ATTN} and DarkTrack2021\cite{rackerMeetsNight}  solely focus on UAV-based night-time scenarios. In contrast our proposed NT-VOT211 is more diverse and generic in terms of number of attributes, bigger in size, and 
more challenging in terms of difficulty index. On our proposed dataset, SOTA tracking algorithms deliver relatively low performances compared to existing datasets. Fig.~\ref{fig:benchlimit} compares the best performance of SOTA on well-known day-time datasets, existing low-light datasets and the proposed night-time NT-VOT211 benchmark. The decrease in the best performance on a dataset shows the increase in the challenge level. The proposed NT-VOT211 is the most challenging benchmark among  all benchmarks included in these experiments.

In addition to proposing a comprehensive night-time benchmark, we also evaluate 43  SOTA VOT methods on 13 existing datasets (Figure \ref{fig:benchlimit}). The aim of this analysis is to facilitate the deployment of VOT algorithms in the field and also to identify potential areas for improvement. The experimental findings highlight the significance of our dataset, which stands out for its low brightness and unique attribute distribution, in driving advancements in algorithmic performance. Additionally, we make the leaderboard and annotation tools publicly accessible. In summary, our key contributions can be summarized as follows:
\begin{figure*}[t]
\centering
\includegraphics[scale=0.30]{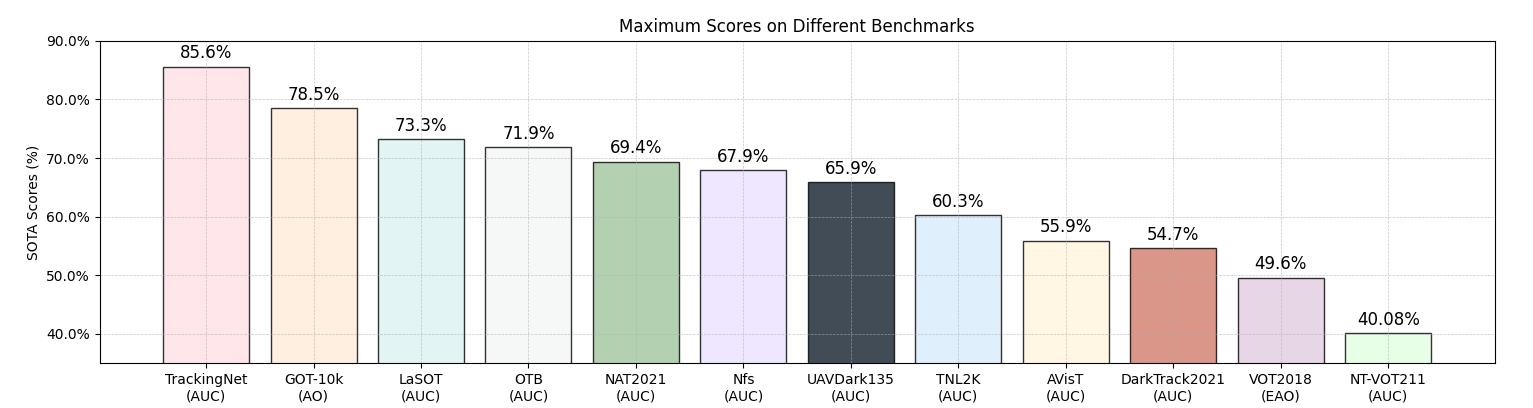}
\caption{The best performing tracking algorithm is shown on each of the 12 benchmarks. AVisT focuses on adverse tracking conditions and UAVDark2021 has majorly low-light scenarios. These are also challenging scenarios for SOTA trackers. NT-VOT211 is unique for night-time content  containing diverse, more challenging and extensive night-time scenes.}

\label{fig:benchlimit}
\end{figure*}
\begin{itemize}
\item \textbf{Comprehensive Night-time Benchmark:} We introduce NT-VOT211, a large-scale benchmark designed for VOT evaluation in night-time conditions. It consists of 211 challenging sequences, totaling  211,000 carefully annotated frames and marked with 8 different attributes. It lies among the large VOT datasets and is notably the \emph{largest dataset} exclusively dedicated to tracking in night-time visibility conditions. Additionally, we make a leaderboard and annotation tools publicly available.

\item \textbf{Large-scale Benchmarking:} We extensively evaluate \emph{43 different trackers} on 13 benchmarks including  NT-VOT211, covering a range of methods, from recent state-of-the-art techniques to foundational approaches like Correlation Filters and Siamese Network based methods (see Figures \ref{yearly_result} and \ref{fig:benchlimit}). 
We provide a thorough analysis of the key distinctions between our proposed benchmark and existing datasets. Based on this analysis, we draw conclusions with insights and recommendations for the field deployment and research directions in night-time VOT.
\end{itemize}

\section{Existing Low-light VOT Benchmarks}
Majority of existing visual object tracking (VOT) benchmarks have focused on tracking well-illuminated scenes while largely ignoring the need for tracker deployment in the low-light conditions or night-time VOT. In the darkness of night, the tracking conditions are much more adverse and pose unique and severe tracking challenges. This may cause inferior tracking performance and even tracking failure. To enable trackers to achieve adequate performance for all day tracking, some researchers have proposed low-light datasets as discussed below.

\textbf{UAVDark135} \cite{UAVDark} dataset was built on the motivation of enabling Unmanned Aerial Vehicles (UAVs) equipped with real-time robust visual trackers for night-time aerial maneuver. 
It comprises of 125K annotated frames in UAV videos.
Along similar lines, \textbf{DarkTrack2021} \cite{rackerMeetsNight} was proposed to facilitate tracking advancements in night-time conditions. It offers 110 challenging sequences with over 100K frames in total. 
Likewise, \textbf{NAT2021} \cite{ATTN} dataset was proposed for unsupervised domain adaptive nighttime tracking. It comprises a test set of 180 manually annotated tracking sequences and a train set of over 276K unlabelled nighttime tracking frames.
To asses tracking performance under adverse visibility conditions, recently \textbf{AVisT} \cite{noman2022avist} dataset was proposed. 
AVisT contains 120 sequences with 80k annotated
frames. The adverse scenarios include heavy rain, dense fog, sandstorm, fire, sun glare, splashing water, low-light, small targets and distractor objects along with camouflage.

Despite some efforts, achieving robust night-time tracking is still a formidable challenge for existing SOTA methods. This is primarily because of a lack of a large-scale night-time benchmark containing diverse and generic night-time scenarios. To this end, we introduce a novel benchmark, NT-VOT211, which comprises diverse content, distinctive low-brightness attributes, and is of large size.
Compared to aforementioned datasets that mainly focus on UAV-based night-time scenarios, our proposed NT-VOT211 offers much bigger diversity in terms of number of attributes; it is more challenging in difficulty index and cause significant performance degradation for existing SOTA. Finally, it is the largest dataset exclusively focused at night-time scenarios.


\section{The Proposed Night-time Benchmark}
In this section, we  compare our proposed night-time benchmark NT-VOT211 with existing benchmarks (sec.~\ref{comp}), we then present  the details of various attributes in night-time benchmark (sec.~\ref{arrsection}), and finally we explain the protocol and  steps used while collecting and annotating the proposed night-time benchmark NT-VOT211 (sec.~\ref{coll_annotate_subsec}).

\subsection{Comparison with other benchmarks}\label{comp}
We compare our benchmark with existing  low-light benchmarks, in terms of illumination level and attribute distribution.

\begin{figure}[t]
	\centering
\includegraphics[scale=0.32]{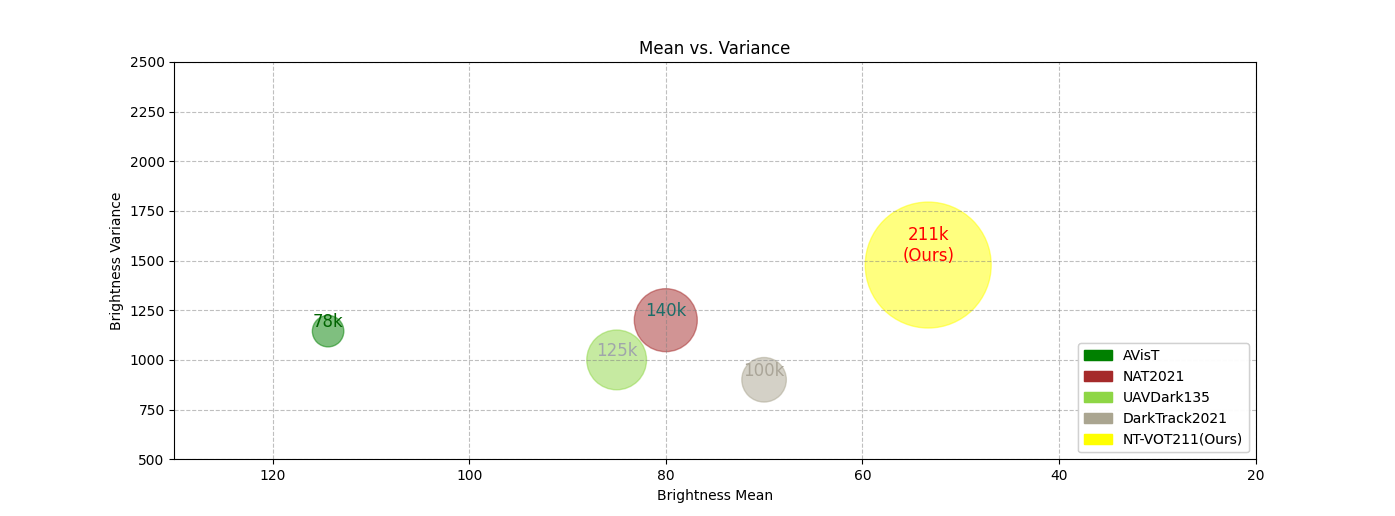}
\caption{Proposed NT-VOT211 benchmark vs. existing low-light datasets for  average brightness (x-axis), brightness variance (y-axis), and the number of total annotated frames (bubble size). The proposed NT-VOT211 benchmark has lowest average intensity, highest brightness variance due to darkness of night and brightness of lights, and has the highest number of annotated frames.   
} 
\label{comparision}
\end{figure}

\noindent\textbf{Illumination Level:} Our statistical analysis considers the mean brightness and brightness variance across five tracking benchmarks including NT-VOT211. The results, shown in Figure \ref{comparision}, indicate that our proposed night-time benchmark (NT-VOT211) has the lowest mean illumination level. It also exhibits the highest brightness variance, highlighting substantial variations in illumination between frames. Furthermore, our dataset stands as the largest benchmark in terms of the number of annotated frames among these benchmarks.

\noindent\textbf{Unified attributes:}
Differing from existing benchmarks where different attributes are labeled by annotators, we introduce a unified criterion including six computer-labeled attributes in addition to two human-labeled attributes. The six computer-labeled attributes can be determined promptly after annotators finish the bounding box annotation, with further details discussed in section \ref{arrsection}. We initially analyze the distribution of these attributes for all mentioned benchmarks (Table \ref{tab1}). It is important to note that, since GOT-10k and TrackingNet did not provide their ground truth, they are not included. Additionally, it's important to note that certain benchmarks lack frame by frame attribute-labels  for occlusion and out-of-view, and this is denoted as "$-$" in the table.

\begin{table*}[ht]
\centering

\caption{Comparison of the percentage of frames per attribute and the difficulty index for  low-light datasets and ours night-time benchmark.}\label{tab1}
\begin{scalebox}{0.65}{
\begin{tabular}{@{}lcccccc@{}}
\toprule
Automated Attributes & {AVisT}\cite{noman2022avist}  &UAVDark135\cite{UAVDark}&NAT2021\cite{ATTN}& DarkTrack2021\cite{rackerMeetsNight} &\textbf{NT-VOT211(Ours)} \\
\midrule
Camera Motion &\textcolor{red}{25.38\%}&4.02\%&3.61\%&6.01\%&8.73\%\\
Deformation &44.46\% &47.55\%&\textcolor{red}{60.49\%} &53.16\%&32.98\%\\
Fast Motion &\textcolor{red}{1.54}\% &1.02\%&0.28\%&0.37\%&1.29\%\\
Motion Blur &20.38\%&18.53\%&14.83\%&\textcolor{red}{29.84\%}&18.21\%\\
Tiny Target &14.38\%&50.15\%&53.50\%&\textcolor{red}{59.43\%}&28.65\%\\
Distractors &29.51\% &6.63\%&28.33\%&24.90\%&\textcolor{red}{39.62\%}\\
\midrule
Manual Attributes\\
\midrule
Occlusion &3.39\% &-&-&-&\textcolor{red}{23.62\%} \\
Out of View &1.42\%&-&-&-&\textcolor{red}{5.21\%} \\
\midrule
Resolution(Average Pixels) &1198x700&1905 x 1072&1280 x 720&2478x1394&1104x818 \\
Difficulty Index &0.34&0.29&0.32&0.36&\textcolor{red}{0.42}\\
\midrule
Contain Training Set &$\times$&$\times$&$\checkmark$&$\times$&$\times$\\
\bottomrule
\end{tabular}}
\end{scalebox}
\end{table*}


\begin{figure*}[h!]
\centering
\includegraphics[width=0.9\textwidth]{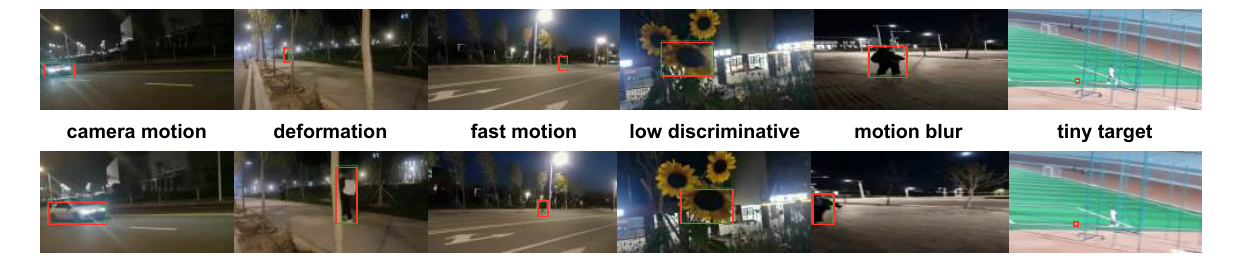}
\caption{Sample frames from our proposed Night-Time Visual Object Tracking (NT-VOT211) dataset with ground truth bounding boxes annotated in \textcolor{red}{red}.}
\label{intro}
\end{figure*}
\subsection{Attributes}\label{arrsection}
NT-VOT211 presents a specialized dataset that encompasses a diverse array of real-world scenarios, mirroring everyday situations where object tracking may occur. Our dataset stands out for its comprehensive coverage of six attributes labeled through automated processes. These attributes comprise camera motion, deformation, fast motion, distractors, motion blur, and tiny target. Beyond these automatically labeled attributes, our dataset incorporates two attributes labeled by human annotators: occlusion and out-of-view. These  attributes are commonplace in other benchmarks as well.
The automated  attributes  are closely related to either the hand-labeled bounding boxes or the properties of the entire frame. In other words, these attributes are generated through automated algorithms that take into account the annotated bounding boxes and the corresponding frame, rather than being subjectively determined by human annotators.

\noindent\textbf{Camera Motion:} Camera motion refers to the detectable motion of the camera during recording. In many benchmarks, camera motion is often overlooked, and it is frequently associated with motion blur\cite{7001050,kiani2017need,fan2019lasot}. However, in our dataset, we have found that camera motion is nearly unavoidable using hand held video recording devices. Particularly in complex scenarios with low brightness (see Figure \ref{intro}), the presence of camera motion can significantly degrade tracking performance. To detect camera motion, we use an optical flow based method by Park et al. \cite{park2004qualitative}. In our dataset \textbf{8.73\%} frames have camera motion (Table \ref{tab1}).

\noindent\textbf{Deformation:} Dealing with deformation poses a common challenge in Visual Object Tracking (VOT) tasks, primarily because the ground truth is often limited to the bounding box of the target object from the initial frame. It is unrealistic to expect the target to maintain its appearance perfectly throughout the entire sequence. In typical VOT tasks, substantial appearance variation is often associated with out-of-plane rotation \cite{7001050}. However, in our dataset, this variation is more likely attributed to factors such as partial occlusion by other objects (as shown in Figure \ref{intro}) or illumination changes, which are more pronounced during night-time. According to the statistics in Table \ref{tab1}, deformation is quite common, in \textbf{32.98\%} of frames in our dataset.

\noindent\textbf{Fast motion:} In the initial implementations of Correlation Filter-based trackers, such as MOSSE \cite{bolme2010visual}, CSK \cite{henriques2012exploiting}, and KCF \cite{henriques2014high}, a fixed searching region was employed for target localization, making them susceptible to challenges posed by fast motion. However, the advent of deep neural networks (DNNs) has significantly mitigated this issue, due to the larger receptive fields. 
Recent research conducted by Vincent et al. \cite{DBLP:conf/eccv/TonkesS22} highlights that multi-head self-attention mechanisms have a tendency to reduce high-frequency signals. Given that fast variations in the image are often associated with high-frequency components, this poses challenges for modern Vision Transformer (ViT)-based trackers like Mixformer \cite{cui2022mixformer}, SeqTrack-L384 \cite{chen2023seqtrack}, and others.
To identify frames with rapid movement, we classify those where the target's displacement between two consecutive frames exceeds 0.5 times the frame's height or width as instances of fast motion. Table \ref{tab1} shows that \textbf{1.29\%} of frames in our dataset exhibit fast motion.

\noindent\textbf{Distractors:} The presence of the distractors attribute indicates that  some  background regions  may have cross-correlation exceeding 0.90 with the target region. Illustrated in Figure \ref{intro}, there is virtually no noticeable distinction between the sunflower within the bounding box and the sunflower in the background. As the camera keeps moving, it's possible for the tracker to occasionally position the bounding box on the sunflower in the background rather than consistently anchoring its prediction on the target. In Table \ref{tab1}, it is evident that our dataset has \emph{a higher proportion of distractors} compared to the other datasets, constituting \textbf{39.62\%} frames.

\noindent\textbf{Motion blur:} Motion blur is widely recognized as a challenging attribute in various benchmarks, such as OTB100\cite{7001050}, NfS \cite{kiani2017need}, and LaSOT\cite{fan2019lasot}. In this dataset, we detect motion blur specifically for the target using the sharpness estimation algorithm proposed by Zhengzi et al. \cite{zhengzi2010fast}. According to Table \ref{tab1}, \textbf{18.21\%} of frames in our dataset exhibit motion blur attributes.

\noindent\textbf{Tiny target:} Tiny target is a condition where the width or height of the target is less than 3.00\% that of the  frame's width or height. As in Figure \ref{intro}, tracking tiny targets is exceptionally challenging, even for human observers, and this challenge is \emph{further amplified in low-light conditions}. Our dataset contains \textbf{28.65\%} frames with tiny targets (Table \ref{tab1}).
   \begin{figure*}[t]
      \centering
      \includegraphics[scale=0.35]{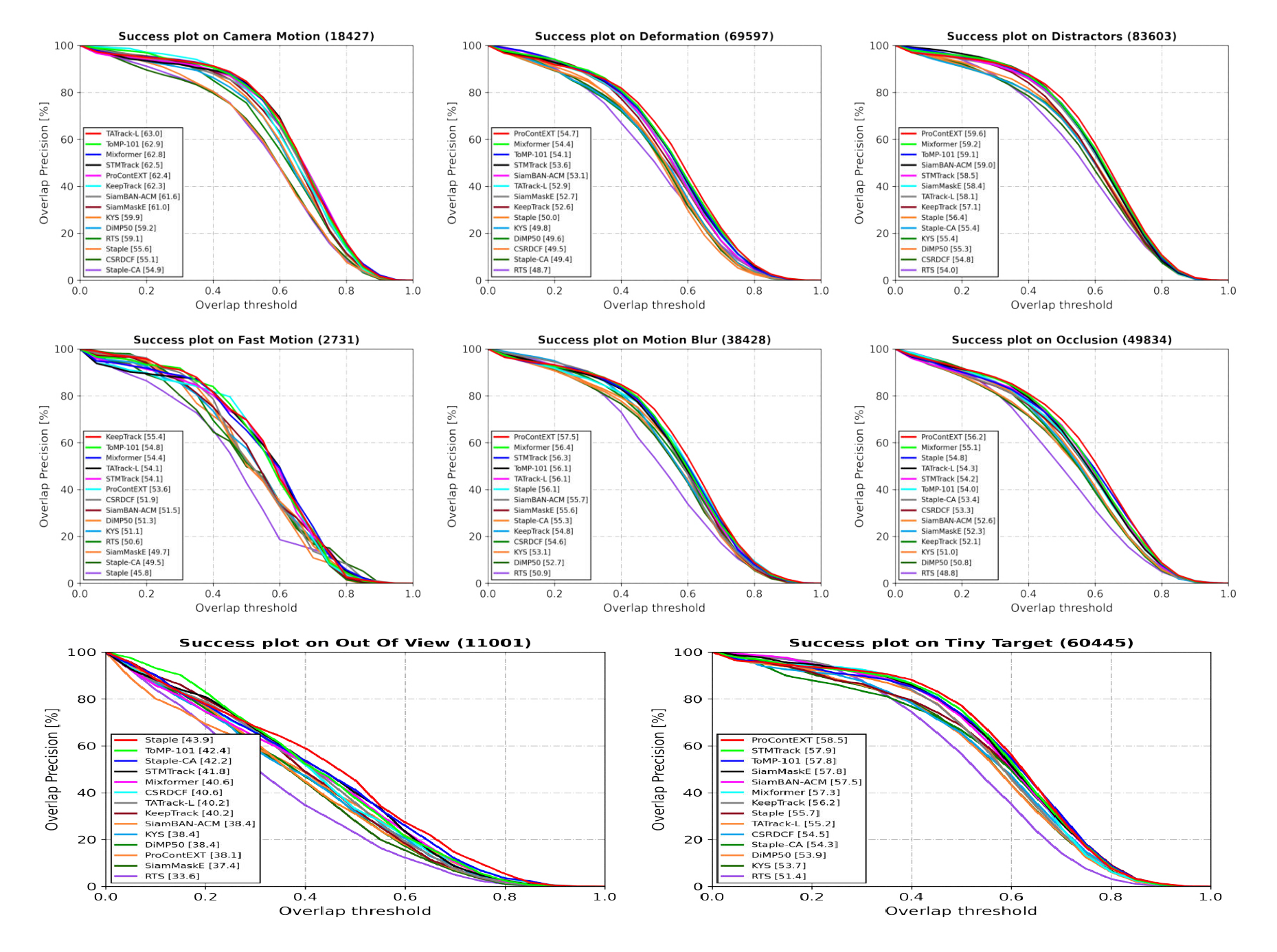}
      \caption{\textbf{Attribute wise Evaluation:} In our evaluation, we consider 3 Correlation Filter trackers, 3 Siamese trackers, 4 Discriminative Correlation Filter (DCF) trackers, and 4 transformer based trackers. Unlike traditional benchmarks that analyze entire video sequences with all frames labeled with a particular attribute, our methodology selectively assesses specific frames with an attribute labels to report a more precise evaluation. The exact count of frames is explicitly specified in the title above the graph. This tailored evaluation approach ensures high precision in gauging tracker performance within our unique context, deviating from the broader assessments conducted by conventional benchmarks. }
      \label{attributedanalysis}
   \end{figure*}
\vspace{-0.5mm}

To conclude this section, we present an evaluation based on different attributes for the top-performing trackers in various categories. Specifically, we select the leading Correlation Filter tracker, Siamese Tracker, and Vision Transformer (ViT)-based tracker from our leaderboard. The outcomes of this assessment are visualized in Figure \ref{attributedanalysis}. The presented results indicate that challenges associated with attributes can be ordered in ascending order of difficulty: camera motion, distractors, tiny target, motion blur, occlusion, fast motion, deformation and out-of-view. We can calculate the Difficulty Index ($D_I$) for all  benchmarks using following formula:
\begin{equation}
D_I=\sum_i p_i \log\left(1 + (1-p_i)(1-s_i)\right),
\end{equation}
where $p_i$ is the percentage of frames with $i$-th attribute and $s_i$ is the maximum score on that attribute. The results are listed in the Table \ref{tab1}, Our method has a difficulty index of 0.42, \emph{the highest among all benchmarks}.

\subsection{Collection and Annotation}
\label{coll_annotate_subsec}
In the  dataset capturing process, 10 persons were involved. Specifically, 5 collectors were assigned to record the video, 3 annotators were utilized in both the coarse and refined annotation phases, and 2 professional examiners conducted the final review of each stage from collection to annotation.

\noindent\textbf{Collection:} All the videos in our dataset were recorded using handheld devices to capture real-world scenarios occurring after sunset. In the interest of ethical and legal considerations, and to respect privacy rights, our data collectors ensured that individuals present during the recordings were informed of the video capture. Their explicit consent was obtained for using this footage in the creation of a publicly available dataset. The result was a collection of 450 videos, comprising an impressive 600,000 frames. To ensure the fundamental quality of the coarse annotation, we employ cross-correlation values to assess the accuracy of the current annotation. If the value falls below 0.7, it signifies a lower quality annotation, and the annotator will need to re-annotate.

Following that, as part of our effort to create a challenging test set for evaluating tracking algorithms, we carefully create a compilation of videos where target tracking presented a substantial challenge, even for human observers. To achieve that, we conducted initial assessments on all 450 coarse annotated sequences utilizing various state-of-the-art tracking algorithms, including, Mixformer\cite{cui2022mixformer} (SOTA on LaSOT\cite{fan2019lasot}), ARTrack-L\cite{wei2023autoregressive} (SOTA on GOT-10k\cite{huang2019got}), SiamMask\_E (SOTA on VOT2018\cite{DBLP:conf/eccv/KristanLMFPZVBL18}), STMTrack\cite{fu2021stmtrack} (SOTA on OTB100\cite{7001050}), NeighborTrack-OSTrack\cite{Chen_2023_CVPR}(SOTA on UAV123\cite{mueller2016benchmark}) and, AiATrack\cite{gao2022aiatrack} (SOTA on NfS\cite{kiani2017need}).
%
To ensure a thorough evaluation of the test set, we initially assessed and computed average performance scores for the mentioned trackers across the entire set. Subsequently, we ranked the videos based on their area-under-the-curve (AUC) scores in descending order. From the top 400 videos, encompassing 500,000 frames, we randomly sampled a subset of 209,560 frames distributed among 210 videos where the target remained consistently trackable. To enhance scenario diversity, we introduced a video capturing a tiny skyborne object at dawn, introducing an additional layer of difficulty due to the target's reduced scale. The finalized dataset comprises 211 videos spanning 211,000 frames, with individual sequence lengths varying from 70 to 10,372 frames. 
\begin{figure}[]
\centering
\includegraphics[scale=0.50]{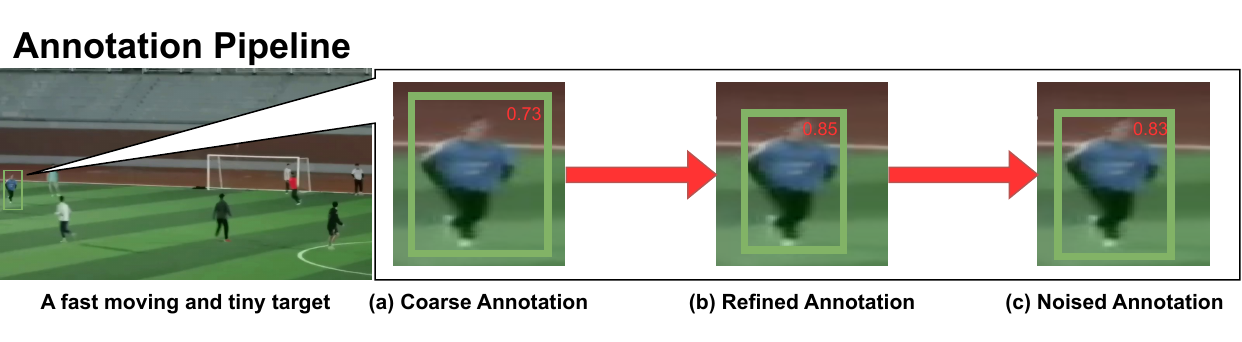}
\caption{A fast moving tiny target in NT-VOT211. Additionally, we mark the cross-correlation values with the previous frame in the top-right corner of each image. Coarse annotation is refined and 5\% annotations are with added noise.}
\label{AnnotationPipeline} \vspace{-1em}
\end{figure}

\noindent\textbf{Annotation:} 
To annotate the video dataset, we developed custom Python software for both coarse and fine bounding box annotation. As mentioned above, coarse annotations were required to select the top 400 videos from the pool of 450 candidates. In the coarse annotation phase, we meticulously annotated only the first frame of each sequence. Subsequently, we efficiently annotate the successive frames, drawing bounding boxes rapidly without imposing strict requirements for perfect centering on the target. To ensure a basic level of quality for the initial assessments of all 450 sequences, we implemented a check based on the normalized cross-correlation between successive frames. If this value dropped below 0.7, the annotator will be prompted to re-examine the current frame before proceeding.
%
%
Following the initial coarse bounding box annotation, we advanced to a fine annotation process involving iterative refinement and careful examination to ensure high-quality results. Three annotators carefully adjusted the coarse boxes utilizing specialized software that provided flexibility in modifying labels as necessary. Subsequently, an examiner conducted a comprehensive review of the refined annotations employing a preview tool, identifying frames requiring additional revision. To avoid potential mistakes, the fine annotations underwent 2 cycles of refinement by annotators, each followed by auditing and feedback from the examiner. This collaborative annotation pipeline, with iterative checks, proved instrumental in upholding stringent labeling quality standards.

\begin{figure*}[!htbp]
\centering
\includegraphics[width=0.75\textwidth]{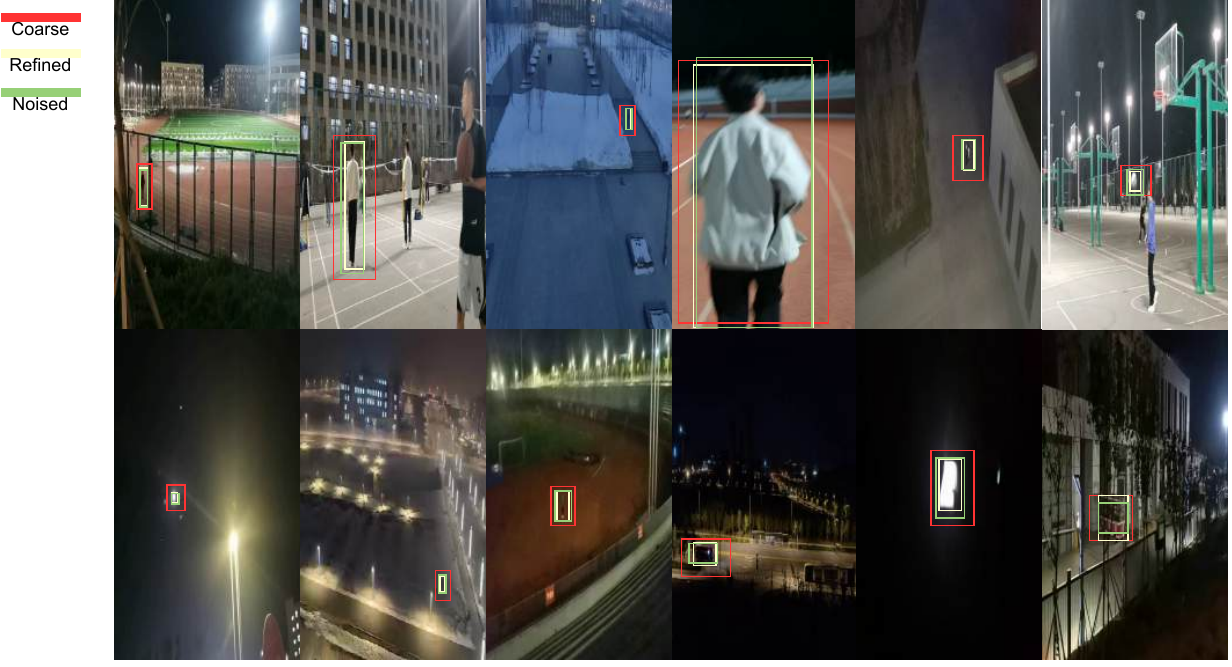}
\caption{Sample image from the NT-VOT211 dataset showing coarse annotation, refined annotation, and noisy annotations for these specific frames. Around 5\% annotations are made noisy.}
\label{quality}
\vspace{-5mm}
\end{figure*}

After that, we intentionally introduced a small degree of label noise to better simulate real-world ambiguity. Approximately 5\% of the box annotations, excluding the first frame, underwent slight random perturbations. Specifically,we applied shifts of $\min\{max\{\frac{[x,y,w,h]}{20},1\},10\}$ pixels from the original annotations. As in the GOT-10k dataset\cite{huang2019got}, we opted not to explicitly identify the perturbed frames, requiring trackers to inherently handle annotation noise during training and evaluation.
We have sampled some examples in Figure \ref{quality} to demonstrate the quality of our annotations.
\vspace{-5mm}

\section{Benchmarking NT-VOT211}

\begin{table*}[t]
\footnotesize
\centering
\caption{Comprehensive Evaluation on our NT-VOT211 benchmark}\label{bench}
\begin{scalebox}{0.55}{
\begin{tabular}{@{}lccccccccc@{}}
\toprule
Method & SOTA & \multicolumn{2}{c}{Tag} & AUC & precision & Norm Precision & OP50 & OP75 & rank(A$|$P) \\
\midrule
ADTrack\cite{li2021adtrack}   & UAVDark70\cite{li2021adtrack}  & \multicolumn{2}{c}{Correlation Filter} & 31.82 & 40.31&68.84&41.31&11.19& 29$|$29       \\
ADTrackV2\cite{UAVDark} & UAVDark135\cite{UAVDark} & \multicolumn{2}{c}{Correlation Filter} & 30.71&39.12&66.21&41.33&10.78 & 35$|$31  \\
AiATrack\cite{gao2022aiatrack} & Nfs\cite{kiani2017need} & \multicolumn{2}{c}{Transformer} &38.91 & 53.47 & 82.42&50.74&12.97& 6$|$6  \\
\multirow{2}{*}[3pt]{ARTrack-L\cite{wei2023autoregressive}} & GOT-10k\cite{huang2019got} & \multicolumn{2}{c}{\multirow{2}{*}[3pt]{Transformer}} &\multirow{2}{*}[3pt]{35.92} & \multirow{2}{*}[3pt]{51.64} & \multirow{2}{*}[3pt]{82.02}&\multirow{2}{*}[3pt]{44.25}&\multirow{2}{*}[3pt]{11.63}& 17$|$13  \\
& TNL2K\cite{wang2021towards} &  &  &   &  & & &  \\
BACF\cite{kiani2017learning} & $\times$ & \multicolumn{2}{c}{Correlation Filter} &26.29 & 35.05 & 62.88&30.96&7.54& 36$|$35  \\
CN\cite{danelljan2014adaptive} & $\times$ & \multicolumn{2}{c}{Correlation Filter} &23.09 & 29.14&55.92&6.76&29.14& 39$|$39  \\
CSK\cite{henriques2012exploiting} & $\times$ & \multicolumn{2}{c}{Correlation Filter} &21.31 & 26.51&54.54&24.19&6.15& 40$|$40  \\
CSRDCF\cite{danelljan2015learning} & $\times$ & \multicolumn{2}{c}{Correlation Filter} &31.59 & 40.31&68.84&40.30&10.26& 29$|$29  \\
DaSiamRPN\cite{zhu2018distractor} & $\times$ & \multicolumn{2}{c}{Siamese} &31.12 & 39.09&65.38&39.76&5.45& 31$|$31  \\
DAT\cite{zhu2018distractor} & $\times$ & \multicolumn{2}{c}{Statistics} &17.52 & 21.60&48.48&19.51&4.49& 42$|$42  \\
DiMP-50\cite{bhat2019learning} & $\times$ & \multicolumn{2}{c}{Discriminative} &35.89 &48.68&77.85&46.31&11.41& 19$|$16  \\
DiMP+STC\cite{rackerMeetsNight} & DarkTrack2021\cite{rackerMeetsNight} & \multicolumn{2}{c}{Discriminative} &35.99 &48.79&77.91&46.71&11.52& 18$|$17  \\
E.T.Tracker\cite{blatter2023efficient}& $\times$ & \multicolumn{2}{c}{Transformer} &34.38 &46.98&77.08&44.33&11.06& 24$|$21  \\
KCF(HOG)\cite{henriques2014high}& $\times$ & \multicolumn{2}{c}{Correlation Filter} &24.10 &32.06&57.56&27.42&6.60& 37$|$37  \\
KeepTrack\cite{mayer2021learning}& $\times$ & \multicolumn{2}{c}{Discriminative} &39.59 &55.50&85.06&50.52&12.83& \textcolor{green}{2}$|$\textcolor{red}{1}   \\
KSY\cite{bhat2020know}& $\times$ & \multicolumn{2}{c}{Discriminative} &36.02 &48.13&77.93&46.81&11.78& 16$|$19 \\
LDES\cite{li2019robust}& $\times$ & \multicolumn{2}{c}{Correlation Filter} &27.72 &35.19&61.42&35.55&9.20& 35$|$34  \\
LightTrack\cite{yan2021lighttrack}& $\times$ & \multicolumn{2}{c}{Siamese} &32.85 &43.65&73.86&41.26&10.43& 27$|$27  \\ 
\multirow{2}{*}[3pt]{Mixformer\cite{cui2022mixformer}}& LaSOT\cite{fan2019lasot}& &\multicolumn{1}{c}{\multirow{3}{*}{Transformer}} &\multirow{3}{*}{39.23} &\multirow{3}{*}{54.20}&\multirow{3}{*}{84.45}&\multirow{3}{*}{51.03}&\multirow{3}{*}{14.39}& \multirow{3}{*}{5$|$\textcolor{blue}{3}}  \\
(ConvMAE)& TrackingNet\cite{muller2018trackingnet}& & &  & & & & &   \\
& AVisT\cite{muller2018trackingnet}& & &  & & & & &   \\
MKCFup\cite{tang2018high}& $\times$ & \multicolumn{2}{c}{Correlation Filter} &28.04 & 34.94&62.17&35.12&8.68& 33$|$36  \\
MOSSE\cite{bolme2010visual}& $\times$ & \multicolumn{2}{c}{Correlation Filter} &19.55 & 25.29&50.88&20.69&5.01& 41$|$41  \\
\multirow{2}{*}[3pt]{Neighbor-}& \multirow{2}{*}{UAV123\cite{mueller2016benchmark}}& &\multicolumn{1}{c}{\multirow{2}{*}{Transformer}} &\multirow{2}{*}{38.32} &\multirow{2}{*}{52.54}&\multirow{2}{*}{83.42}&\multirow{2}{*}{50.02}&\multirow{2}{*}{14.11}& \multirow{2}{*}{9$|$9}  \\
Track(OSTrack)\cite{Chen_2023_CVPR}& & & &  & & & & &   \\
Ocean\cite{zhang2020ocean}& $\times$ & \multicolumn{2}{c}{Siamese} &32.86 &46.72&77.56&41.07&9.24& 26$|$22  \\
OSTrack-384\cite{ye2022joint}& $\times$ & \multicolumn{2}{c}{Transformer} &38.59 &53.06&83.47&50.41&14.11& 8$|$7  \\
ProContEXT\cite{lan2023procontext}& $\times$ & \multicolumn{2}{c}{Transformer} &40.08 &54.54&84.95&53.25&15.05& \textcolor{red}{1}$|$\textcolor{green}{2}  \\
RTS\cite{paul2022robust}& $\times$ & \multicolumn{2}{c}{Discriminative} &36.20 &53.68&82.58&42.75&10.44& 15$|$5  \\
SiamBAN-ACM\cite{han2020learning}& $\times$ & \multicolumn{2}{c}{Siamese} &35.80 &48.31&79.01&47.43&12.85& 19$|$18  \\
SiamDW\cite{zhang2019deeper}& $\times$ & \multicolumn{2}{c}{Siamese} &35.18 &46.18&75.92&46.79&13.02& 21$|$25  \\
SiamFC\cite{bertinetto2016fully}& $\times$ & \multicolumn{2}{c}{Siamese} &32.62 &40.81&69.77&42.34&11.45& 28$|$28  \\
SiamMask\cite{wang2019fast}& $\times$ & \multicolumn{2}{c}{Siamese} &35.14 &46.49&77.35&46.92&12.71& 22$|$24  \\
SiamMask\_E& VOT2018\cite{DBLP:conf/eccv/KristanLMFPZVBL18} & \multicolumn{2}{c}{Siamese} &35.22 &46.57&77.42&47.01&12.74& 20$|$23  \\
SiamRPN\cite{li2018high} & $\times$ & \multicolumn{2}{c}{Siamese} &33.92 &44.04&74.27&44.58&11.77& 25$|$26  \\
SLT-TransT\cite{kim2022towards} & $\times$ & \multicolumn{2}{c}{Transformer} &37.22 &51.70&82.55&47.85&12.96& 11$|$12  \\
Staple\cite{Bertinetto_2016_CVPR} & $\times$ & \multicolumn{2}{c}{Correlation Filter} &31.29 &39.12&67.68&40.73& 10.71& 30$|$30  \\
Staple-CA\cite{mueller2017context} & $\times$ & \multicolumn{2}{c}{Correlation Filter} &30.68 &38.57&66.40&39.83& 10.43& 32$|$32  \\
STARK\cite{yan2021learning} & $\times$ & \multicolumn{2}{c}{Transformer} &38.26 &51.37&81.11&51.06&13.66& 10$|$14  \\
STMTrack\cite{fu2021stmtrack} & OTB100\cite{7001050} & \multicolumn{2}{c}{Siamese} &36.84 &50.34&80.87&48.12&13.94& 12$|$16  \\
STRCF\cite{li2018learning}& $\times$ & \multicolumn{2}{c}{Correlation Filter} &27.86 &36.18&62.69&34.18&7.64& 34$|$33  \\
TATrack-L\cite{he2023target} &$\times$ & \multicolumn{2}{c}{Transformer} &39.29 &53.94&84.84&51.77&14.21& \textcolor{blue}{3}$|$4  \\
ToMP-50\cite{mayer2022transforming} &$\times$ & \multicolumn{2}{c}{Transformer} &39.25 &53.01&82.83&51.98&14.32& 4$|$8   \\
ToMP-101\cite{mayer2022transforming} &$\times$ & \multicolumn{2}{c}{Transformer} &38.61 &51.98&81.93&51.08&14.29& 7$|$10  \\
TransT\cite{chen2021transformer} &$\times$ &\multicolumn{2}{c}{Transformer} &36.79&51.97&82.43&46.71&11.87& 13$|$11 \\
TRAS\cite{Dunnhofer_2020_ACCV}&$\times$ &\multicolumn{2}{c}{Discriminative} &23.58&30.64&61.26&27.16&6.85& 38$|$38  \\
TrDiMP\cite{wang2021transformer}&$\times$ &\multicolumn{2}{c}{Transformer} &36.66&50.68&80.30&46.04&11.38& 14$|$15  \\
UDAT\cite{ATTN}  & NAT2021\cite{ATTN}    & \multicolumn{2}{c}{Transformer}            & 33.99 &46.51&75.12&42.21&10.10& 23$|$21       \\
Unicorn\cite{yan2022towards}&$\times$ &\multicolumn{2}{c}{Transformer} &34.52&47.77&76.06&43.60&11.10& 23$|$20 \\

\bottomrule
\end{tabular}}
\end{scalebox}
\end{table*}
%

We evaluated NT-VOT211 by benchmarking it against 43 state-of-the-art trackers gathered from 42 papers. These trackers were implemented in various codebases such as PySot, Video Analyst, and PyTracking. To facilitate this evaluation, we created specific dataloaders for each of these three toolkits, and the corresponding code is made publicly available. To ensure a fair comparison, all trackers were run in their original environments, producing raw results on NT-VOT211.
Following the execution, predictions were post-processed and assessed using standardized metrics within PyTracking. These metrics include AUC\cite{fan2019lasot}, precision\cite{7001050}, normalized precision\cite{kiani2017need}, OP50, and OP75\cite{huang2019got}. The detailed results can be found in Table \ref{bench}, where "Rank(A$|$P)" indicates the global rank of each method based on its performance in terms of AUC$|$Precision.
\subsection{Analysis}
\noindent\textbf{Correlation Filter Trackers:} MOSSE\cite{bolme2010visual} represents a seminal work in the field of correlation filter tracking. Interestingly, it is the lowest-ranking method in our benchmark, which is a statistics-based approach. This result is understandable because in many sequences, the background consists of purely black pixels with a value of 0, which can impede statistical processing. A similar issue is observed with Staple-CA\cite{mueller2017context}.
Conversely, we observe significant improvement with methods like KCF(HOG)\cite{henriques2014high} and CN\cite{danelljan2014adaptive}, which indicates the potential of feature diversity. Staple\cite{Bertinetto_2016_CVPR} argues that combining the final score map of correlation filter (CF) methods with the appearance model generated by DAT\cite{zhu2018distractor} leads to substantial enhancements. MKCFup\cite{tang2018high} suggests that considering multiple results from different CF trackers simultaneously can be beneficial.
BACF \cite{kiani2017learning} adopts a different approach by focusing on the target itself rather than modeling the background, resulting in improved success. On the other hand, STRCF \cite{li2018learning} and CSRDCF \cite{danelljan2015learning} emphasize spatio-temporal constraints, which have proven to be effective for tracking in low-light conditions.

\noindent\textbf{Discriminative Trackers:}  The DCF trackers ensure feature diversity through pretrained backbones.  KSY \cite{bhat2020know}  leverages information across different frames to distinguish foreground from background, achieving a notable 16-th rank in AUC on NT-VOT211. 
KeepTrack \cite{mayer2021learning}  demonstrates that gradient-descent-based methods are quite advanced compared to plain appearance models.
RTS\cite{paul2022robust} attempts to fuse the score map with segmentation results in a more complex manner. However, it turns out to be less efficient than KSY, possibly due to the challenges of segmentation posed by low brightness conditions.

\noindent\textbf{Siamese Trackers:} SiamFC\cite{bertinetto2016fully}  primarily focused on the target itself. SiamRPN\cite{li2018high} achieves better performance. Surprisingly, the appearance-model-based DaSiamRPN\cite{zhu2018distractor} does not outperform its baseline SiamRPN on NT-VOT211. SiamMask\cite{wang2019fast} and its advanced iteration SiamMask\_E\cite{DBLP:conf/eccv/KristanLMFPZVBL18} have integrated a binary segmentation objective into their loss functions, thereby achieving enhanced tracking accuracy.
SiamDW\cite{zhang2019deeper} investigates the depth of the backbone network to balance performance and processing speed, although the improvement on NT-VOT211 is limited. SiamBAN-ACM\cite{han2020learning} introduces Asymmetric Convolution to accommodate different feature map sizes, proved to be effective for nighttime tracking. Ocean\cite{zhang2020ocean} opts for direct prediction of the position and scale of target objects in an anchor-free manner. Interestingly, it yields degraded results on NT-VOT211, suggesting that traditional Siamese methods may have advantages. This could be due to the challenges posed by the high amount of random noise in nighttime frames when estimating the position instead of anchor offsets.
STMTrack~\cite{fu2021stmtrack} concentrates on the target itself and achieves a 13-th rank on AUC in NT-VOT211. LightTrack~\cite{yan2021lighttrack} introduces an efficient approach to developing a lightweight and effective tracker. However, it falls behind SiamFC in frames per second (FPS), and its performance improvement on our dataset is marginal compared to SiamFC, which might be explained by the unique distribution pattern compared to other benchmarks.

\noindent\textbf{Transformer-based Trackers:} TransT\cite{chen2021transformer} is one of the earliest transformer-based tracking methods. SLT-TransT\cite{kim2022towards} recognizes visual tracking as a sequence-level task and introduces a sequence-level training strategy, ranking 11-th in AUC on NT-VOT211. STARK\cite{yan2021learning} focuses on learning spatio-temporal transformers for visual tracking and achieves a 10-th ranking  in AUC on NT-VOT211. TrDiMP\cite{wang2021transformer} is ranked 14-th by incorporating temporal context.
Mixformer\cite{cui2022mixformer} proposes a Mixed Attention Module to replace the standard Siamese architecture's classification and regression branches, benefiting from the contributions of the masked convolution autoencoders (ConvMAE) module\cite{wu2021cvt}. AiATrack\cite{gao2022aiatrack} introduces efficient feature reuse and target-background embeddings to make full use of temporal references, achieving the 6-th rank in NT-VOT211. OSTrack\cite{ye2022joint} adopts a multi-feature fusing strategy, ranking 8-th on NT-VOT211. However, based on OSTrack, NeighborTrack-OSTrack\cite{Chen_2023_CVPR} models the region around the target and experiences a slight decrease in performance on NT-VOT211.
ToMP-50\cite{mayer2022transforming}, utilizing the ResNet-50\cite{he2016deep} backbone, extends the model predictor to estimate a second set of weights that are applied for accurate bounding box regression, similar to the Siamese approach. It achieves a rank of 4 in AUC on NT-VOT211. Interestingly, ToMP-101 with the ResNet-101 backbone performs less effectively, suggesting that deeper backbone is not necessarily better for NT-VOT211. ARTrack-L\cite{wei2023autoregressive} conducts regression on position and ranks 17-th in AUC. E.T.Tracker\cite{blatter2023efficient} is a lightweight tracker, while Unicorn\cite{yan2022towards} is a multi-task classifier using the same network with the same parameters. ProContEXT\cite{lan2023procontext}, leverages both temporal and spatial contexts, instead of sole reliance on the target area in the initial frame, significantly enhancing tracking precision and achieve SOTA performance achieved 1-st rank on NT-VOT211.

\begin{figure}[t]
  \begin{minipage}[!htbp]{.39\linewidth}
\scalebox{0.75}{
\begin{tabular}{lcl}
\hline
ProContEXT                              & \multicolumn{1}{l}{Original} & Fine-tuned \\ \hline
\multicolumn{1}{l|}{\textbf{Attributed Scores}}    & \multicolumn{2}{c}{AUC scores}     \\ \hline
\multicolumn{1}{l|}{camera motion} & 37.1 & {\color[HTML]{CB0000} 39.4} \\
\multicolumn{1}{l|}{deformation}   & 54.2 & {\color[HTML]{CB0000} 55.2} \\
\multicolumn{1}{l|}{fast motion}   & 31.4 & {\color[HTML]{CB0000} 34.1} \\
\multicolumn{1}{l|}{motion blur}   & 47.5 & {\color[HTML]{CB0000} 49.1} \\
\multicolumn{1}{l|}{tiny target}   & 29.1 & {\color[HTML]{CB0000} 29.7} \\
\multicolumn{1}{l|}{distractors} & 60.7                                                     & {\color[HTML]{CB0000} 65.8} \\ \hline
\multicolumn{1}{l|}{\textbf{Overall Scores}} & 51.2                                                     & {\color[HTML]{CB0000} 52.3} \\ \hline
\end{tabular}} \vspace{-1em}
 \captionof{table}{ Original results (no finetuning) vs. finetuned res on Split Training Sequences in NT-VOT211.}
\label{tab:fine-tuning_results}
  \end{minipage}\hfill
  \begin{minipage}[!htbp]{.55\linewidth}

 \includegraphics[scale=0.2]{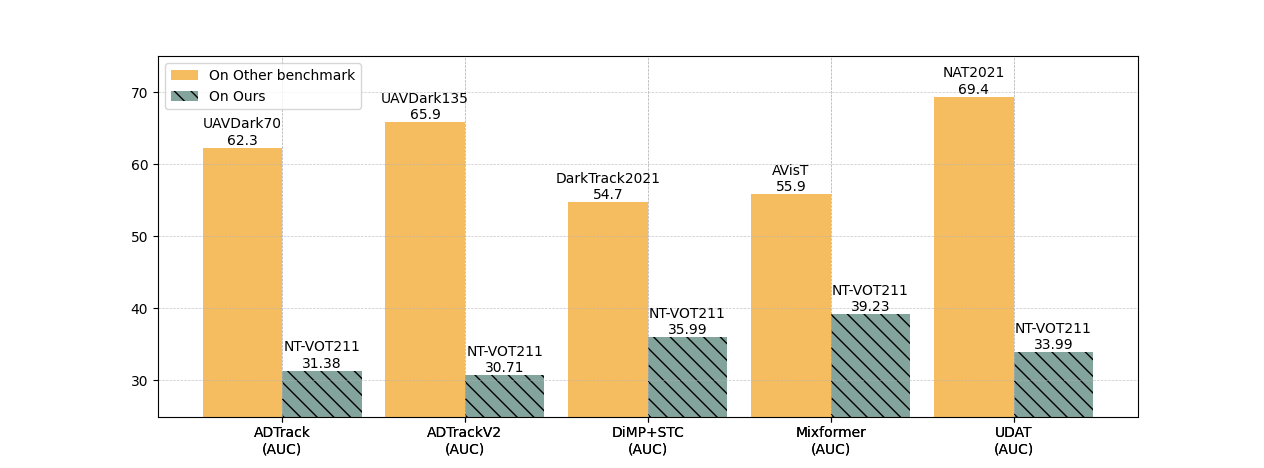}
    \caption{Trackers particularly developed for night-time scenarios that record the best performance on existing night-time datasets showcase significantly poor performance on ours NT-VOT211 benchmark.}
    \label{fig:best_performing_low_light_tracker}  
  \vspace{-5mm}
   
  \end{minipage}
\end{figure}

\noindent\textbf{Fine-tuning performance:} To demonstrate the impact of fine-tuning in NT-VOT211, we extract sequences with over 20\% of frames containing specific attributes as test sequences. We created test sequences for camera motion, deformation, fast motion, motion blur, tiny targets, and distractors features. We use the remaining sequences as the training set (test: training 37\%:63\%). Table \textcolor{red}{\ref{tab:fine-tuning_results}} shows the comparison of original results with the fine tuned for ProContEXT tracker. Fine-tuning results in marginal improvements compared to no finetuning, thereby revealing the pressing need for new tracking algorithms capable of handling diverse night-time scenarios.




\noindent\textbf{Night-time trackers on NT-VOT211:} We observe the performance of trackers specifically targeted at night-time scenarios and that reveal the best performance on existing night-time datasets tend to perform poorly on ours NT-VOT211 benchmark (see Figure~\ref{fig:best_performing_low_light_tracker}).
\vspace{-4mm}
\subsection{Insights for Future Research}



\noindent\textbf{Focus on the target:} Trackers like DAT~\cite{zhu2018distractor}, Staple-CA~\cite{mueller2017context}, DaSiamRPN~\cite{zhu2018distractor}, and NeighborTrack-OSTrack~\cite{Chen_2023_CVPR} exhibit degraded performance under low illumination due to over-reliance on background context. Aggressively extracting information from the surroundings hinders these methods in dim settings with less discriminative context. On the other hand, target-focused trackers like CN~\cite{danelljan2014adaptive}, KCF~\cite{henriques2014high}, and SiamRPN~\cite{li2018high} prove more robust in low light. While background context remains valuable, as demonstrated by KeepTrack's architecture \cite{mayer2021learning}, which distinguishes the target from the background, achieving optimal tracking robustness requires balancing target-centric and contextual reasoning, especially in handling illumination variation. Target-only approaches excel in invariance but lack discrimination, while context-driven methods enable classification at the cost of sensitivity to environmental changes.


\noindent\textbf{Spatio-temporal information is critical:} Methods that incorporate spatio-temporal priors\cite{yan2021learning,danelljan2015learning,lan2023procontext,wang2021transformer,gao2022aiatrack} consistently outperform their baselines, emphasizing the significance of motion and temporal consistency in low-light tracking. By leveraging target dynamics and coherence across frames, trackers extract discriminative cues without depending on rich visual appearance in low light settings. The reliability of frame-to-frame variations in inferring spatio-temporal patterns persists even under changes in illumination. These findings underscore the importance of encoding target movement and temporal consistency for achieving invariance to environmental conditions.

\noindent\textbf{Transformer based Trackers are better:} The fact that 9 of the top 10 trackers on NT-VOT211 utilize a vision transformer architecture highlights the benefits of this approach for generalization. The inherent ability of transformers to model long-range dependencies appears well-suited for tracking under challenging new conditions not seen during training. 

\noindent\textbf{Regression on offset:} Regression on offset is necessary, as evidenced by the success of SOTA ProContEXT\cite{lan2023procontext} and TATrack\cite{he2023target}. However, the failure of Ocean\cite{zhang2020ocean} and the performance gap between ARTrack\cite{wei2023autoregressive} and ProContEXT suggest that directly regressing on location may not be a suitable choice.

\noindent\textbf{The deeper may not the better:} The poor performance of ToMP-101\cite{mayer2022transforming} and SiamDW\cite{zhang2019deeper} on the proposed dataset suggests that these methods may be overfitting to their training datasets and may not generalize well to the proposed NT-VOT211 dataset.

\vspace{-3mm}
\section{Conclusion}
In this work, a novel night-time visual object tracking benchmark,  NT-VOT211, is proposed. It is a new large-scale benchmark for night-time visual tracking, featuring 211 challenging videos with 211K well-annotated frames. Comprehensive evaluation and analysis indicates the complementary nature of this benchmark to the existing ones for evaluation and deployment of the trackers in real-world systems. The state-of-the-art performance on NT-VOT211 highlights significant room for improvement in the task of nighttime visual object tracking. Besides, we provide a leaderboard, showcasing performance rankings, and annotation tools. 
%
%
%

\begin{thebibliography}{8}
\bibitem{Bertinetto_2016_CVPR}
Bertinetto, L., Valmadre, J., Golodetz, S., Miksik, O., Torr, P.H.S.: Staple: Complementary learners for real-time tracking. In: Proceedings of the IEEE Conference on Computer Vision and Pattern Recognition (CVPR) (June 2016)

\bibitem{bertinetto2016fully}
Bertinetto, L., Valmadre, J., Henriques, J.F., Vedaldi, A., Torr, P.H.: Fully-convolutional siamese networks for object tracking. In: Computer Vision--ECCV 2016 Workshops: Amsterdam, The Netherlands, October 8-10 and 15-16, 2016, Proceedings, Part II 14. pp. 850--865. Springer (2016)

\bibitem{bhat2019learning}
Bhat, G., Danelljan, M., Gool, L.V., Timofte, R.: Learning discriminative model prediction for tracking. In: Proceedings of the IEEE/CVF international conference on computer vision. pp. 6182--6191 (2019)

\bibitem{bhat2020know}
Bhat, G., Danelljan, M., Van~Gool, L., Timofte, R.: Know your surroundings: Exploiting scene information for object tracking. In: Computer Vision--ECCV 2020: 16th European Conference, Glasgow, UK, August 23--28, 2020, Proceedings, Part XXIII 16. pp. 205--221. Springer (2020)

\bibitem{bhat2020learning}
Bhat, G., Lawin, F.J., Danelljan, M., Robinson, A., Felsberg, M., Van~Gool, L., Timofte, R.: Learning what to learn for video object segmentation. In: Computer Vision--ECCV 2020: 16th European Conference, Glasgow, UK, August 23--28, 2020, Proceedings, Part II 16. pp. 777--794. Springer (2020)

\bibitem{blatter2023efficient}
Blatter, P., Kanakis, M., Danelljan, M., Van~Gool, L.: Efficient visual tracking with exemplar transformers. In: Proceedings of the IEEE/CVF Winter Conference on Applications of Computer Vision. pp. 1571--1581 (2023)

\bibitem{bolme2010visual}
Bolme, D.S., Beveridge, J.R., Draper, B.A., Lui, Y.M.: Visual object tracking using adaptive correlation filters. In: 2010 IEEE computer society conference on computer vision and pattern recognition. pp. 2544--2550. IEEE (2010)

\bibitem{chen2023seqtrack}
Chen, X., Peng, H., Wang, D., Lu, H., Hu, H.: Seqtrack: Sequence to sequence learning for visual object tracking. In: Proceedings of the IEEE/CVF Conference on Computer Vision and Pattern Recognition. pp. 14572--14581 (2023)

\bibitem{chen2021transformer}
Chen, X., Yan, B., Zhu, J., Wang, D., Yang, X., Lu, H.: Transformer tracking. In: Proceedings of the IEEE/CVF conference on computer vision and pattern recognition. pp. 8126--8135 (2021)

\bibitem{Chen_2023_CVPR}
Chen, Y.H., Wang, C.Y., Yang, C.Y., Chang, H.S., Lin, Y.L., Chuang, Y.Y., Liao, H.Y.M.: Neighbortrack: Single object tracking by bipartite matching with neighbor tracklets and its applications to sports. In: Proceedings of the IEEE/CVF Conference on Computer Vision and Pattern Recognition (CVPR) Workshops. pp. 5138--5147 (June 2023)

\bibitem{cui2022mixformer}
Cui, Y., Jiang, C., Wang, L., Wu, G.: Mixformer: End-to-end tracking with iterative mixed attention. In: Proceedings of the IEEE/CVF Conference on Computer Vision and Pattern Recognition. pp. 13608--13618 (2022)

\bibitem{danelljan2015learning}
Danelljan, M., Hager, G., Shahbaz~Khan, F., Felsberg, M.: Learning spatially regularized correlation filters for visual tracking. In: Proceedings of the IEEE international conference on computer vision. pp. 4310--4318 (2015)

\bibitem{danelljan2014adaptive}
Danelljan, M., Shahbaz~Khan, F., Felsberg, M., Van~de Weijer, J.: Adaptive color attributes for real-time visual tracking. In: Proceedings of the IEEE conference on computer vision and pattern recognition. pp. 1090--1097 (2014)

\bibitem{Dunnhofer_2020_ACCV}
Dunnhofer, M., Martinel, N., Micheloni, C.: Tracking-by-trackers with a distilled and reinforced model. In: Proceedings of the Asian Conference on Computer Vision (ACCV) (November 2020)

\bibitem{fan2019lasot}
Fan, H., Lin, L., Yang, F., Chu, P., Deng, G., Yu, S., Bai, H., Xu, Y., Liao, C., Ling, H.: Lasot: A high-quality benchmark for large-scale single object tracking. In: Proceedings of the IEEE/CVF conference on computer vision and pattern recognition. pp. 5374--5383 (2019)

\bibitem{fu2021stmtrack}
Fu, Z., Liu, Q., Fu, Z., Wang, Y.: Stmtrack: Template-free visual tracking with space-time memory networks. In: Proceedings of the IEEE/CVF conference on computer vision and pattern recognition. pp. 13774--13783 (2021)

\bibitem{fukushima1980self}
Fukushima, K.: A self-organizing neural network model for a mechanism of pattern recognition unaffected by shift in position. Biol, Cybern  \textbf{36},  193--202 (1980)

\bibitem{gao2022aiatrack}
Gao, S., Zhou, C., Ma, C., Wang, X., Yuan, J.: Aiatrack: Attention in attention for transformer visual tracking. In: European Conference on Computer Vision. pp. 146--164. Springer (2022)

\bibitem{han2020learning}
Han, W., Dong, X., Khan, F.S., Shao, L., Shen, J.: Learning to fuse asymmetric feature maps in siamese trackers. In: {IEEE} Conference on Computer Vision and Pattern Recognition, {CVPR} 2021, virtual, June 19-25, 2021. pp. 16570--16580. Computer Vision Foundation / {IEEE} (2021). \doi{10.1109/CVPR46437.2021.01630}

\bibitem{he2023target}
He, K., Zhang, C., Xie, S., Li, Z., Wang, Z.: Target-aware tracking with long-term context attention. In: Williams, B., Chen, Y., Neville, J. (eds.) Thirty-Seventh {AAAI} Conference on Artificial Intelligence, {AAAI} 2023, Thirty-Fifth Conference on Innovative Applications of Artificial Intelligence, {IAAI} 2023, Thirteenth Symposium on Educational Advances in Artificial Intelligence, {EAAI} 2023, Washington, DC, USA, February 7-14, 2023. pp. 773--780. {AAAI} Press (2023). \doi{10.1609/AAAI.V37I1.25155}, \url{https://doi.org/10.1609/aaai.v37i1.25155}

\bibitem{he2016deep}
He, K., Zhang, X., Ren, S., Sun, J.: Deep residual learning for image recognition. In: Proceedings of the IEEE conference on computer vision and pattern recognition. pp. 770--778 (2016)

\bibitem{henriques2012exploiting}
Henriques, J.F., Caseiro, R., Martins, P., Batista, J.: Exploiting the circulant structure of tracking-by-detection with kernels. In: Computer Vision--ECCV 2012: 12th European Conference on Computer Vision, Florence, Italy, October 7-13, 2012, Proceedings, Part IV 12. pp. 702--715. Springer (2012)

\bibitem{henriques2014high}
Henriques, J.F., Caseiro, R., Martins, P., Batista, J.: High-speed tracking with kernelized correlation filters. IEEE transactions on pattern analysis and machine intelligence  \textbf{37}(3),  583--596 (2014)

\bibitem{huang2019got}
Huang, L., Zhao, X., Huang, K.: Got-10k: A large high-diversity benchmark for generic object tracking in the wild. IEEE transactions on pattern analysis and machine intelligence  \textbf{43}(5),  1562--1577 (2019)

\bibitem{kiani2017need}
Kiani~Galoogahi, H., Fagg, A., Huang, C., Ramanan, D., Lucey, S.: Need for speed: A benchmark for higher frame rate object tracking. In: Proceedings of the IEEE International Conference on Computer Vision. pp. 1125--1134 (2017)

\bibitem{kiani2017learning}
Kiani~Galoogahi, H., Fagg, A., Lucey, S.: Learning background-aware correlation filters for visual tracking. In: Proceedings of the IEEE international conference on computer vision. pp. 1135--1143 (2017)

\bibitem{kim2022towards}
Kim, M., Lee, S., Ok, J., Han, B., Cho, M.: Towards sequence-level training for visual tracking. In: European Conference on Computer Vision. pp. 534--551. Springer (2022)

\bibitem{DBLP:conf/eccv/KristanLMFPZVBL18}
Kristan, M., Leonardis, A., Matas, J., Felsberg, M., Pflugfelder, R.P., Zajc, L.C., Voj{\'{\i}}r, T., Bhat, G., Lukezic, A., Eldesokey, A., Fern{\'{a}}ndez, G., Garc{\'{\i}}a{-}Mart{\'{\i}}n, {\'{A}}., Iglesias{-}Arias, {\'{A}}., Alatan, A.A., Gonz{\'{a}}lez{-}Garc{\'{\i}}a, A., Petrosino, A., Memarmoghadam, A., Vedaldi, A., Muhic, A., He, A., Smeulders, A.W.M., Perera, A.G., Li, B., Chen, B., Kim, C., Xu, C., Xiong, C., Tian, C., Luo, C., Sun, C., Hao, C., Kim, D., Mishra, D., Chen, D., Wang, D., Wee, D., Gavves, E., Gundogdu, E., Velasco{-}Salido, E., Khan, F.S., Yang, F., Zhao, F., Li, F., Battistone, F., Ath, G.D., Subrahmanyam, G.R.K.S., Bastos, G.S., Ling, H., Galoogahi, H.K., Lee, H., Li, H., Zhao, H., Fan, H., Zhang, H., Possegger, H., Li, H., Lu, H., Zhi, H., Li, H., Lee, H., Chang, H.J., Drummond, I., Valmadre, J., Martin, J.S., Chahl, J.S., Choi, J.Y., Li, J., Wang, J., Qi, J., Sung, J., Johnander, J., Henriques, J.F., Choi, J., van~de Weijer, J., Herranz, J.R., Mart{\'{\i}}nez, J.M., Kittler, J.,
  Zhuang, J., Gao, J., Grm, K., Zhang, L., Wang, L., Yang, L., Rout, L., Si, L., Bertinetto, L., Chu, L., Che, M., Maresca, M.E., Danelljan, M., Yang, M., Abdelpakey, M.H., Shehata, M.S., Kang, M., Lee, N., Wang, N., Miksik, O., Moallem, P., Vicente{-}Mo{\~{n}}ivar, P., Senna, P., Li, P., Torr, P.H.S., Raju, P.M., Qian, R., Wang, Q., Zhou, Q., Guo, Q., Nieto, R.M., Gorthi, R.K.S.S., Tao, R., Bowden, R., Everson, R.M., Wang, R., Yun, S., Choi, S., Vivas, S., Bai, S., Huang, S., Wu, S., Hadfield, S., Wang, S., Golodetz, S., Tang, M., Xu, T., Zhang, T., Fischer, T., Santopietro, V., Struc, V., Wang, W., Zuo, W., Feng, W., Wu, W., Zou, W., Hu, W., Zhou, W., Zeng, W., Zhang, X., Wu, X., Wu, X., Tian, X., Li, Y., Lu, Y., Law, Y.W., Wu, Y., Demiris, Y., Yang, Y., Jiao, Y., Li, Y., Zhang, Y., Sun, Y., Zhang, Z., Zhu, Z., Feng, Z., Wang, Z., He, Z.: The sixth visual object tracking {VOT2018} challenge results. In: Leal{-}Taix{\'{e}}, L., Roth, S. (eds.) Computer Vision - {ECCV} 2018 Workshops - Munich, Germany,
  September 8-14, 2018, Proceedings, Part {I}. Lecture Notes in Computer Science, vol. 11129, pp. 3--53. Springer (2018). \doi{10.1007/978-3-030-11009-3\_1}, \url{https://doi.org/10.1007/978-3-030-11009-3\_1}

\bibitem{VOT_TPAMI}
Kristan, M., Matas, J., Leonardis, A., Vojir, T., Pflugfelder, R., Fernandez, G., Nebehay, G., Porikli, F., \v{C}ehovin, L.: A novel performance evaluation methodology for single-target trackers. IEEE Transactions on Pattern Analysis and Machine Intelligence  \textbf{38}(11),  2137--2155 (Nov 2016). \doi{10.1109/TPAMI.2016.2516982}

\bibitem{lan2023procontext}
Lan, J.P., Cheng, Z.Q., He, J.Y., Li, C., Luo, B., Bao, X., Xiang, W., Geng, Y., Xie, X.: Procontext: Exploring progressive context transformer for tracking. In: ICASSP 2023-2023 IEEE International Conference on Acoustics, Speech and Signal Processing (ICASSP). pp.~1--5. IEEE (2023)

\bibitem{li2018high}
Li, B., Yan, J., Wu, W., Zhu, Z., Hu, X.: High performance visual tracking with siamese region proposal network. In: Proceedings of the IEEE conference on computer vision and pattern recognition. pp. 8971--8980 (2018)

\bibitem{li2021adtrack}
Li, B., Fu, C., Ding, F., Ye, J., Lin, F.: Adtrack: Target-aware dual filter learning for real-time anti-dark uav tracking. In: 2021 IEEE international conference on robotics and automation (ICRA). pp. 496--502. IEEE (2021)

\bibitem{UAVDark}
Li, B., Fu, C., Ding, F., Ye, J., Lin, F.: All-day object tracking for unmanned aerial vehicle. {IEEE} Trans. Mob. Comput.  \textbf{22}(8),  4515--4529 (2023). \doi{10.1109/TMC.2022.3162892}, \url{https://doi.org/10.1109/TMC.2022.3162892}

\bibitem{li2018learning}
Li, F., Tian, C., Zuo, W., Zhang, L., Yang, M.H.: Learning spatial-temporal regularized correlation filters for visual tracking. In: Proceedings of the IEEE conference on computer vision and pattern recognition. pp. 4904--4913 (2018)

\bibitem{li2019robust}
Li, Y., Zhu, J., Hoi, S.C., Song, W., Wang, Z., Liu, H.: Robust estimation of similarity transformation for visual object tracking. In: Proceedings of the AAAI conference on artificial intelligence. vol.~33, pp. 8666--8673 (2019)

\bibitem{mayer2022transforming}
Mayer, C., Danelljan, M., Bhat, G., Paul, M., Paudel, D.P., Yu, F., Van~Gool, L.: Transforming model prediction for tracking. In: Proceedings of the IEEE/CVF conference on computer vision and pattern recognition. pp. 8731--8740 (2022)

\bibitem{mayer2021learning}
Mayer, C., Danelljan, M., Paudel, D.P., Van~Gool, L.: Learning target candidate association to keep track of what not to track. In: Proceedings of the IEEE/CVF International Conference on Computer Vision. pp. 13444--13454 (2021)

\bibitem{mueller2016benchmark}
Mueller, M., Smith, N., Ghanem, B.: A benchmark and simulator for uav tracking. In: Computer Vision--ECCV 2016: 14th European Conference, Amsterdam, The Netherlands, October 11--14, 2016, Proceedings, Part I 14. pp. 445--461. Springer (2016)

\bibitem{mueller2017context}
Mueller, M., Smith, N., Ghanem, B.: Context-aware correlation filter tracking. In: Proceedings of the IEEE conference on computer vision and pattern recognition. pp. 1396--1404 (2017)

\bibitem{muller2018trackingnet}
Muller, M., Bibi, A., Giancola, S., Alsubaihi, S., Ghanem, B.: Trackingnet: A large-scale dataset and benchmark for object tracking in the wild. In: Proceedings of the European conference on computer vision (ECCV). pp. 300--317 (2018)

\bibitem{noman2022avist}
Noman, M., Ghallabi, W.A., Kareem, D., Mayer, C., Dudhane, A., Danelljan, M., Cholakkal, H., Khan, S., Gool, L.V., Khan, F.S.: Avist: {A} benchmark for visual object tracking in adverse visibility. In: 33rd British Machine Vision Conference 2022, {BMVC} 2022, London, UK, November 21-24, 2022. p.~817. {BMVA} Press (2022), \url{https://bmvc2022.mpi-inf.mpg.de/817/}

\bibitem{park2004qualitative}
Park, S.C., Lee, H.S., Lee, S.W.: Qualitative estimation of camera motion parameters from the linear composition of optical flow. Pattern Recognition  \textbf{37}(4),  767--779 (2004)

\bibitem{paul2022robust}
Paul, M., Danelljan, M., Mayer, C., Van~Gool, L.: Robust visual tracking by segmentation. In: European Conference on Computer Vision. pp. 571--588. Springer (2022)

\bibitem{tang2018high}
Tang, M., Yu, B., Zhang, F., Wang, J.: High-speed tracking with multi-kernel correlation filters. In: Proceedings of the IEEE conference on computer vision and pattern recognition. pp. 4874--4883 (2018)

\bibitem{DBLP:conf/eccv/TonkesS22}
Tonkes, V., Sabatelli, M.: How well do vision transformers (vts) transfer to the non-natural image domain? an empirical study involving art classification. In: Karlinsky, L., Michaeli, T., Nishino, K. (eds.) Computer Vision - {ECCV} 2022 Workshops - Tel Aviv, Israel, October 23-27, 2022, Proceedings, Part {I}. Lecture Notes in Computer Science, vol. 13801, pp. 234--250. Springer (2022). \doi{10.1007/978-3-031-25056-9\_16}, \url{https://doi.org/10.1007/978-3-031-25056-9\_16}

\bibitem{wang2021transformer}
Wang, N., Zhou, W., Wang, J., Li, H.: Transformer meets tracker: Exploiting temporal context for robust visual tracking. In: Proceedings of the IEEE/CVF conference on computer vision and pattern recognition. pp. 1571--1580 (2021)

\bibitem{wang2019fast}
Wang, Q., Zhang, L., Bertinetto, L., Hu, W., Torr, P.H.: Fast online object tracking and segmentation: A unifying approach. In: Proceedings of the IEEE/CVF conference on Computer Vision and Pattern Recognition. pp. 1328--1338 (2019)

\bibitem{wang2021towards}
Wang, X., Shu, X., Zhang, Z., Jiang, B., Wang, Y., Tian, Y., Wu, F.: Towards more flexible and accurate object tracking with natural language: Algorithms and benchmark. In: Proceedings of the IEEE/CVF Conference on Computer Vision and Pattern Recognition. pp. 13763--13773 (2021)

\bibitem{wei2023autoregressive}
Wei, X., Bai, Y., Zheng, Y., Shi, D., Gong, Y.: Autoregressive visual tracking. In: Proceedings of the IEEE/CVF Conference on Computer Vision and Pattern Recognition. pp. 9697--9706 (2023)

\bibitem{wu2021cvt}
Wu, H., Xiao, B., Codella, N., Liu, M., Dai, X., Yuan, L., Zhang, L.: Cvt: Introducing convolutions to vision transformers. In: Proceedings of the IEEE/CVF international conference on computer vision. pp. 22--31 (2021)

\bibitem{7001050}
Wu, Y., Lim, J., Yang, M.H.: Object tracking benchmark. IEEE Transactions on Pattern Analysis and Machine Intelligence  \textbf{37}(9),  1834--1848 (2015). \doi{10.1109/TPAMI.2014.2388226}

\bibitem{yan2022towards}
Yan, B., Jiang, Y., Sun, P., Wang, D., Yuan, Z., Luo, P., Lu, H.: Towards grand unification of object tracking. In: European Conference on Computer Vision. pp. 733--751. Springer (2022)

\bibitem{yan2021learning}
Yan, B., Peng, H., Fu, J., Wang, D., Lu, H.: Learning spatio-temporal transformer for visual tracking. In: Proceedings of the IEEE/CVF international conference on computer vision. pp. 10448--10457 (2021)

\bibitem{yan2021lighttrack}
Yan, B., Peng, H., Wu, K., Wang, D., Fu, J., Lu, H.: Lighttrack: Finding lightweight neural networks for object tracking via one-shot architecture search. In: Proceedings of the IEEE/CVF Conference on Computer Vision and Pattern Recognition. pp. 15180--15189 (2021)

\bibitem{ye2022joint}
Ye, B., Chang, H., Ma, B., Shan, S., Chen, X.: Joint feature learning and relation modeling for tracking: A one-stream framework. In: European Conference on Computer Vision. pp. 341--357. Springer (2022)

\bibitem{rackerMeetsNight}
Ye, J., Fu, C., Cao, Z., An, S., Zheng, G., Li, B.: Tracker meets night: {A} transformer enhancer for {UAV} tracking. {IEEE} Robotics Autom. Lett.  \textbf{7}(2),  3866--3873 (2022). \doi{10.1109/LRA.2022.3146911}, \url{https://doi.org/10.1109/LRA.2022.3146911}

\bibitem{ATTN}
Ye, J., Fu, C., Zheng, G., Paudel, D.P., Chen, G.: Unsupervised domain adaptation for nighttime aerial tracking. In: {IEEE/CVF} Conference on Computer Vision and Pattern Recognition, {CVPR} 2022, New Orleans, LA, USA, June 18-24, 2022. pp. 8886--8895. {IEEE} (2022). \doi{10.1109/CVPR52688.2022.00869}, \url{https://doi.org/10.1109/CVPR52688.2022.00869}

\bibitem{zhang2019deeper}
Zhang, Z., Peng, H.: Deeper and wider siamese networks for real-time visual tracking. In: Proceedings of the IEEE/CVF conference on computer vision and pattern recognition. pp. 4591--4600 (2019)

\bibitem{zhang2020ocean}
Zhang, Z., Peng, H., Fu, J., Li, B., Hu, W.: Ocean: Object-aware anchor-free tracking. In: Computer Vision--ECCV 2020: 16th European Conference, Glasgow, UK, August 23--28, 2020, Proceedings, Part XXI 16. pp. 771--787. Springer (2020)

\bibitem{zhengzi2010fast}
Zhengzi, W., Zhihua, X., Cuiqun, H.: A fast quality assessment of image blur based on sharpness. In: 2010 3rd International Congress on Image and Signal Processing. vol.~5, pp. 2302--2306. IEEE (2010)

\bibitem{zhu2018distractor}
Zhu, Z., Wang, Q., Li, B., Wu, W., Yan, J., Hu, W.: Distractor-aware siamese networks for visual object tracking. In: Proceedings of the European conference on computer vision (ECCV). pp. 101--117 (2018)

\end{thebibliography}
%

\end{document}